\documentclass[fleqn,10pt]{wlscirep}
\usepackage[utf8]{inputenc}
\usepackage[T1]{fontenc}
\usepackage{graphicx}
\usepackage{tikz}
\usetikzlibrary{positioning}

\title{Multi-scale fMRI time series analysis for understanding neurodegeneration in MCI}

\author[1]{Ammu R.}
\author[1,*]{Debanjali Bhattacharya}
\author[2]{Ameiy Acharya}
\author[2]{Ninad Aithal}
\author[2]{Neelam Sinha}

\affil[1]{International Institute of Information Technology (IIIT) Bangalore, Bangalore 560100, India}
\affil[2]{Centre for brain research, Indian Institute of Science (IISc), Bangalore 560012, India}

\affil[*]{debanjali.bhattacharya@iiitb.ac.in, jhimli20@gmail.com}

\begin{abstract}

In this study, we present a technique that spans multi-scale views (global scale- meaning brain network-level and local scale- examining each individual ROI that constitutes the network) applied to resting-state fMRI volumes. Deep learning based classification is utilized in understanding neurodegeneration. The novelty of the proposed approach lies in utilizing two extreme scales of analysis. One branch considers the entire network within graph-analysis framework. Concurrently, the second branch scrutinizes each ROI within a network independently, focusing on evolution of dynamics. For each subject, graph-based approach employs partial correlation to profile the subject in a single graph where each ROI is a node, providing insights into differences in levels of participation. In contrast, non-linear analysis employs recurrence plots to profile a subject as a multichannel 2D image, revealing distinctions in underlying dynamics. The proposed approach is employed for classification of a cohort of 50 healthy control (HC) and 50 Mild Cognitive Impairment (MCI), sourced from ADNI dataset. 
Results point to: (1) reduced activity in ROIs such as PCC in MCI (2) greater activity in occipital in MCI, which is not seen in HC (3) when analysed for dynamics, all ROIs in MCI show greater predictability in time-series.

\end{abstract}
\begin{document}

\flushbottom
\maketitle



\section*{Introduction}
\label{sec:intro}
Mild Cognitive Impairment (MCI) manifests as a discernible decline in cognitive capacities, encompassing memory, language, problem-solving, and thinking skills. The notable difference between MCI and typical age-related cognitive changes is the pronounced decline in cognitive abilities, surpassing what would typically be anticipated based on an individual's age and level of education. Remarkably, this decline, while significant, does not severely impede their day-to-day activities and independence. MCI is often viewed as a transitional phase between the natural cognitive shifts associated with aging and the more severe cognitive deterioration linked to Alzheimer’s disease (AD) or other forms of dementia \cite{anderson2019state,petersen1995normal}. While not everyone with MCI progresses to dementia, it is recognized as a potential precursor or early stage of AD in certain cases \cite{janoutova2015mild}.Nevertheless, it's crucial to emphasize that not all individuals with MCI develop AD or any other form of dementia. Identifying individuals with MCI early on is of paramount importance as it provides a window for timely diagnosis and intervention \cite{morley2011anticholinergic}. Detecting MCI in its nascent stages empowers healthcare professionals to initiate suitable treatments and interventions, potentially slowing down the progression of cognitive decline and mitigating symptoms. Furthermore, early diagnosis enables individuals and their families to plan for the future, make essential lifestyle adjustments, and access necessary support services.
A comprehensive examination of alterations in brain networks and the activation of distinct brain regions grants us invaluable insights into the intricate interaction between these regions. MCI, characterized by cognitive impairment, profoundly affects numerous brain networks, giving rise to challenges across diverse cognitive domains \cite{refa}. 
The impact of MCI on brain networks is significant, giving rise to disruptions that resonate throughout the cognitive domain. These disruptions become crucial focal points for understanding the condition and devising interventions to mitigate cognitive decline in those affected by MCI.
One of the prominently impacted brain networks in individuals with MCI is the default mode network (DMN), a network that becomes active during periods of rest or when the mind is not engaged in any specific task. Individuals with MCI exhibit a noticeable decrease in connectivity and functional abnormalities within the DMN \cite{refc,ammu1}. These alterations within the DMN may serve as early indicators of AD, given that changes in this network are frequently intertwined with the progression of the disease.
Additionally, MCI can disrupt other significant brain networks, such as the frontoparietal network, which plays a pivotal role in executive functions like planning, problem-solving, and decision-making. Disruptions in the frontoparietal network due to MCI can lead to challenges in these cognitive processes. Moreover, the occipital lobe, primarily associated with visual processing, forms the occipital network, crucial for tasks related to visual perception, object recognition, and spatial processing. The sensory networks are specialized in processing sensory input from various modalities, including vision, hearing, touch, taste, and smell, each involving the respective sensory cortices in the brain. The cingulate cortex plays a central role in the cingulate network, contributing to emotional regulation, pain perception, and decision-making. The cerebellum, once believed to be primarily involved in motor control, is now recognized for its role in various brain networks, contributing to motor coordination, cognition, and emotional processing. 
Understanding these nuanced alterations within various brain networks not only provides critical insights into the mechanisms underlying MCI but also opens up new avenues for potential diagnostic and therapeutic strategies. The potential for early detection and targeted interventions holds promise in ultimately delaying or ameliorating the cognitive decline associated with MCI and related neurodegenerative conditions.\par
The present study utilizes fMRI time series of Dosenbach ROIs which are selected from six classical brain networks: the default mode network (34 ROIs), fronto-parietal (21 ROIs), cingulo-opercular (32 ROIs), sensorimotor (33 ROIs), occipital (22 ROIs), and cerebellum (18 ROIs). Each of these 160 Dosenbach ROIs is represented as a time series, and we focus our investigation in analyzing these time series data.
The most natural approach to analyze fMRI time series is to build a graph, motivating our graph-based methodology. In this representation, a brain network is modeled as a graph where each node encapsulates the time series information for a specific ROI, and two nodes are connected by an edge if their corresponding time series exhibit correlation. This global perspective provides insights into the activity levels of ROIs within a network. Several reported works have also utilised non linear analysis to understand the dynamics. Some methods convert time series into images for the analysis of underlying dynamics. Popular techniques include Gramian Angular Fields (GAF), Markov Transition Fields (MTF) \cite{gafmtf1,gafmtf2}, cellular automata, etc., which transform time series into images based on diverse properties such as alternate coordinate systems and transition probabilities. However, a tool that leverages non-linear analysis is the Recurrence Plot visualization, aiming to comprehend the dynamics driving the time series. A. Lombardi et al. \cite{complexsys}, performed non-linear analysis on fMRI data to compare spatio-temporal properties of BOLD signal for schizophrenia detection. The authors in literature \cite{complexsys} used recurrence plots and then extracts features from the recurrence quantification analysis (RQA). Another work of A. Lombardi et al. uses cross recurrence plots \cite{7145165} and extracts the RQA measures and creates a graph based out of these measures for schizophrenia detection. However, no existing work combines multiple scale analysis for holistic inference. In the proposed work, we study activity of ROIs at brain network level as well as individual ROIs to understand their corresponding dynamics. We profile each subject into six network-specific graphs and multichannel recurrence plots. The detailed methodology is outlined in Proposed methodology section, which serves as the basis for the classification between HC and MCI.

\section*{Results}
\label{sec:result}
\subsection*{Dataset Description}
\label{sec:datasetdescrp}
\subsubsection*{Subjects}
We examined fMRI images of 50 people diagnosed with Mild Cognitive Impairment (MCI) and another 50 individuals without any cognitive issues, all gathered from the Alzheimer’s Disease Neuroimaging Initiative (ADNI) database \cite{adni1}.  The individuals with MCI displayed no additional neurodegenerative conditions apart from MCI, whereas the cognitively healthy group had no history of cognitive decline, stroke, or noteworthy psychiatric ailments. 

\subsubsection*{Image Acquisition and Pre-processing}
\label{sec:preprocessing}
3T MRI scanner from Siemens was utilized for acquiring the MRI scans. Initially, anatomical images in T1-MPRAGE sequence were obtained, echo time (TE = 3 ms), repetition time (TR = 2300 ms), flip angle (9°), and voxel size (1 × 1 × 1 mm). Subsequently, resting-state fMRI (RS-fMRI) scans were acquired using the same 3T Siemens MRI scanner, employing an echo-planar imaging sequence. The acquisition parameters for RS-fMRI scans included echotime (TE) of 30 ms, repetition time (TR) of 3000 ms, 197 slices, slice thickness of 3.4 mm, flip angle of 90 degrees, and pixel spacing of 3.4 mm.

The data were processed using FSL (FMRIB Software Library) version 6.0.4 \cite{jenkinson2012fsl}, which was made available by the Centre for Functional MRI of the Brain. Processing of fMRI images involved an initial step of motion correction, adjustment for slice timing, normalization to MNI space, and regression to account for nuisance variables.

For the raw T1 images, the automated Brain Extraction Tool within FSL was utilized to remove non-brain tissue. The functional images at lower resolution were aligned to the structural images using a linear registration approach with 6 degrees of freedom. To enable inter-subject comparisons, the MRI slices were standardized to a predefined template \cite{ref30} through spatial normalization. Further, the functional data were aligned with the Montreal Neurological Institute (MNI) space. Tissue segmentation and linear registration were carried out using FSL tools. To address the T1-equilibration effect, the initial ten volumes of the functional image were excluded. Additionally, steps were taken to correct for motion and slice timing, as well as to strip the skull. Mitigating the impact of head motion involved regressing out motion parameters using a regression model that incorporated 24 motion regressors. Nuisance variables, such as white matter and cerebrospinal fluid regressors, were also included. The resulting residuals were then used for subsequent processing.

   
\subsection*{Local scale: ROI-specific analysis using Recurrence plots}
\label{}
Recurrence plots being symmetric in nature can have various patterns which explains the evolution of the underlying dynamical system. The most prominent patterns in recurrence plots are the diagonal and the lines parallel to it on which most of the traditional RQA measures are based.  Horizontal lines with their corresponding vertical lines give raise to Box like structure, Stationary processes produces smooth laminar areas, unstructured signals can give raise to rugged tire-like patterns in recurrence plots. The next sub-sections describe certain characteristics of recurrence plots which affirms our hypothesis that human brain can be conceptualised as a non-linear dynamical system and MCI subjects have a structured dynamiocs but healthy controls lack thereof, which also enables good classification between HC and MCI. Table~\ref{tab:Attributes_RP} summarizes the attributes of recurrence plots.\par

\textbf{Diagonal-parallel lines:}
\label{sec:diagonal}
The diagonals in recurrence plots are important since they convey information about the evolution of states over time, and the presence of a pattern or lack thereof. \cite{RecurrencePlotWebsite, marwan2007recurrence}. For example, longer lines parallel to the diagonal implies higher stability and higher predictability for a long time. Thicker lines parallel to diagonal reinforce this attribute, showing a sub-sequence of neighbouring states are evolving in a similar manner. In HC, shorter and thinner lines parallel to diagonal are observed. This indicates relatively lesser structure and lesser predictability. Dark diagonal-parallel lines indicate that the sub-sequence of states are closer to each other; while bright diagonal-parallel lines indicate that the sub-sequence of states are far from each other. In HC, bright diagonal-parallel lines are observed, while in MCI dark diagonal-parallel lines are observed. \par

\textbf{Horizontal-vertical lines:}
\label{sec:horizontal}
Vertical lines can appear in recurrence plot, suggesting that some states either do not change or change slowly with time, indicating laminar states. Since recurrence plots are symmetrical, these vertical lines will also have corresponding horizontal lines, which appear as a box-like structure. If these lines happen to be bright, it would mean that the state is an outlier in the phase space, and all other states are located far apart. Similarly if these lines happen to be dark, then it indicates that the states are very similar and points to the chances of being at the center of the phase space as it is located in close proximity to every other point in the phase space. These box-like structures are more sharper in HC as compared to MCI.

\begin{table}[ht]
\centering
\begin{tabular}{|p{5cm}|p{5cm}|p{5cm}|}
\hline
\textbf{Attributes} & \textbf{Interpretation in the phase state} & \textbf{Interpretation in dynamics} \\
\hline
White Vertical / Horizontal line & The state is an outlier & Unstable state \\
\hline
Black Vertical / Horizontal line & The state is close to every other state in phase space, indicating the possibility of being at the center & Stable state \\
\hline
Lines parallel to diagonal & The states are evolving in a determined nature & Predictability \\
\hline
Box-like structure & Some states do not change or change slowly for some time & Laminar states \\
\hline
Homogeneity & Closely spaced points in phase space & Stationary \\
\hline
Heterogeneity & Scattered points in phase space & Non-stationary \\
\hline
\end{tabular}
\caption{Attributes observed in Recurrence plots.}
\label{tab:Attributes_RP}
\end{table}

\begin{figure}[ht]
\centering
    \begin{subfigure}{0.224\textwidth}
        \centering
        \includegraphics[width=0.8\linewidth]{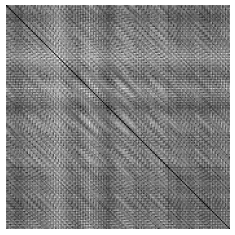}
         \caption{Healthy control}
    \end{subfigure}
    \begin{subfigure}{0.224\textwidth}
        \centering
        \includegraphics[width=0.8\linewidth]{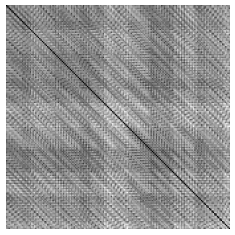}
         \caption{Healthy control}
    \end{subfigure}
    \begin{subfigure}{0.224\textwidth}
        \centering
        \includegraphics[width=0.8\linewidth]{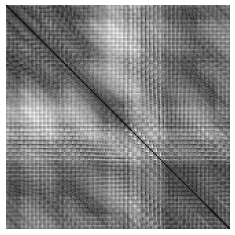}
        \caption{MCI}
    \end{subfigure}
        \begin{subfigure}{0.224\textwidth}
        \centering
        \includegraphics[width=0.8\linewidth]{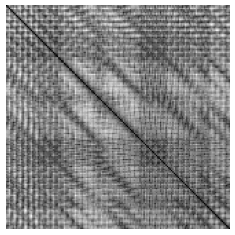}
         \caption{MCI}
    \end{subfigure}
\caption{Recurrence plots illustrate the dynamic evolution of ROI 18 within the Default Mode Network, identified as the most active ROI using a Graph-based community detection approach. The analysis reveals a significant difference in activity in the post cingulate 108 region of the brain between subjects with MCI and healthy individuals.}
\label{fig:samplerecplot}
\end{figure}


\begin{table}[h]
\centering
\begin{tabular}{|l|c|c|c|c|c|c|}
\hline
\textbf{Network} & \textbf{Precision}& \textbf{Recall}& \textbf{F1 Score} & \textbf{Accuracy}\\
\hline
\textbf{Default mode} & 0.74 & 0.74 & 0.74 & 77.78\%\\
\textbf{Frontoparietal} & 0.75 & 0.75 & 0.75 & 78.95\%\\
\textbf{Occipital} & 0.91 & 0.91 & 0.91& 89.47\%\\
\textbf{Sensorimotor} & 0.80 & 0.89 & 0.84 & 84.21\%\\
\textbf{Cingulo-Opercular} & 0.75 & 0.86 & 0.80 & 84.21\%\\
\textbf{Cerebellum} & 0.70 & 0.88 & 0.81 & 80.00\%\\
\hline
\end{tabular}
\caption{Validation accuracy in classifying subjects as healthy and MCI using CNN based DNN model on Recurrence plots.}
\label{tab:Results_RP}
\end{table}

\begin{table}[h]
\centering
\begin{tabular}{|l|c|c|c|c||c|c|c|c|c|}
\hline

\multirow{2}{*}{\textbf{Network}} & \multicolumn{4}{c||}{\textbf{Eigenvalues}} & \multicolumn{4}{c|}{\textbf{Eigenvectors}} \\
\cline{2-9}
  & \textbf{Precision} & \textbf{Recall} & \textbf{F1 score} & \textbf{Accuracy} & \textbf{Precision} & \textbf{Recall} & \textbf{F1 score} & \textbf{Accuracy}\\
\hline
\textbf{Default}  & 0.84 & 0.88 & 0.86 & 84\% & 0.94 & 0.87 & 0.90 & 89\%  \\
\textbf{Frontoparietal}  & 0.92 & 0.81 & 0.86 & 87\%  & 0.83 & 0.91 & 0.87 & 87\%\\
\textbf{Occipital}  & 0.84 & 0.87 & 0.85 & 86\%  & 0.86 & 0.88 & 0.87 & 86\%\\
\textbf{Sensorimotor}  & 0.86 & 0.87 & 0.86 & 86\% & 0.85 & 0.89 & 0.87 & 87\%\\
\textbf{Cingulo-Opercular}  & 0.85 & 0.82 & 0.83 & 85\% & 0.86 & 0.88 & 0.87 & 86\%\\
\textbf{Cerebellum}  & 0.86 & 0.82 & 0.84 & 83\% & 0.84 & 0.85 & 0.84 & 84\%\\
\hline
\end{tabular}
\caption{Performance measures results using Ensemble classifier to categorize HC Vs. MCI}
\label{tab:tableameiy}
\end{table}


\subsection*{Global scale: brain network analysis using graphs}

We have conducted two types of feature-based classification
with Eigen values and Eigen vector corresponding to the highest Eigen value, calculated from fMRI time series of six different brain networks. Table~\ref{tab:tableameiy} displays the results of the proposed Eigen decomposition based classification. It is seen that Eigen vector corresponding to the Eigen value of DMN resulted in maximum accuracy of 89\% using ensemble classfier. It is seen that the results of this analysis outperforms the results of previous works as reported in literature \cite{comp5,comp6,comp17,comp18}. 
\par
Furthermore, for all subjects, we utilize degree matrices to track node connections in DMN. For this, nodes ordered based on maximum connectivity from the degree matrix which is followed by ranking the ROIs based on their degree of connectivity to every other ROIs, across all subjects. The ranking of top 5 ROIs is shown in Table~\ref{tab:roi}. By doing so, we highlight the most significant nodes shared across subjects, offering insight into their importance and how they are affected by neurodegeneration.
While comparing these top 5 ROIs across all subjects, reduced connectivity in post cingulate is seen in case of MCI, across all MCI subjects as compared to HC. As seen from Table~\ref{tab:roi}, PCC occurs only once in top 5 ROIs (post cingulate 111) than healthy individuals where it appears fours times in top 5 ROIs (post cingulate 108, 111, 73, 93). The PCC is implicated in episodic memory, the ability to remember specific events or experiences. Changes in the PCC may affect memory processes, and memory impairment is a common symptom of MCI. This finding is in line with several studies reported in literature \cite{scheff2015synaptic,PCC,PCC1} where the authors identified posterior cingulate (PCC) region as one of the important ROI which could serve as a potential predictive biomarkers in detecting MCI. Moreover, it is known that in MCI, there is change in activity and functional compensation observed in occipital nodes to offset the loss of functionality in other ROIs\cite{occipital}. Our analysis supports this finding, demonstrating that the ROIs associated with the occipital region exhibit the highest level of activity in individuals with MCI.

\begin{table}[h]
\centering
\begin{tabular}{|c|c|c|c|c|c|c|}
\hline
\multicolumn{3}{|c|}{\textbf{Healthy}} & \multicolumn{3}{|c|}{\textbf{MCI}} \\
\hline
\textbf{ROI no.} & \textbf{Dosenbach ROI} & \textbf{Fraction of subj.} & \textbf{ROI no.} & \textbf{Dosenbach ROI} & \textbf{Fraction of subj.} \\
\hline
18 & post cingulate 108 & 14/50 & 14 & occipital 137 & 19/50 \\
01 & ACC 14 & 13/50 & 13 & occipital 136 & 13/50 \\
09 & inf temporal 63 & 11/50 & 06 & fusiform 84 & 12/50 \\
17 & post cingulate 111 & 11/50 & 09 & inf temporal 63 & 12/50 \\
15 & occipital 92 & 11/50 & 34 & vmPFC 1 & 12/50 \\
03 & aPFC 5 & 10/50 & 32 & vmPFC 7 & 11/50 \\
21 & post cingulate 73 & 10/50 & 17 & post cingulate 111 & 11/50 \\
19 & post cingulate 93 & 10/50 & 31 & vmPFC 11 & 11/50 \\
13 & occipital 136 & 09/50 & 22 & precuneus 132 & 10/50 \\
06 & fusiform 84 & 09/50 & 15 & occipital 92 & 10/50 \\
\hline
\end{tabular}
\caption{Identification of Dosenbach ROIs of DMN that appear at top based on degree of connectivity (ranked from highest to lowest), across all HC and MCI subjects}
\label{tab:roi}
\end{table}

\section*{Discussion}

In the pursuit of early detection and classification of MCI, a critical intermediary stage between normal cognitive aging and more severe conditions like Alzheimer's disease, several research endeavors have ventured into the realm of advanced imaging techniques, machine learning, and deep learning methodologies. A key focus has been structural imaging, particularly the analysis of MRI data since this technique offers a valuable means of examining substantial alterations in brain structure associated with these conditions \cite{ref21}. By quantifying specific changes in these regions, we gain important insights into disease progression and underlying pathology. Perez-Gonzalez et al\cite{perez2021mild} introduced a Random Forest classifier aimed at distinguishing individuals with MCI from healthy control subjects. Their approach seamlessly integrated multimodal data, combining features extracted from structural T1-weighted and diffusion-weighted MRI with neuro-psychological scores. This multifaceted analysis yielded an impressive area under the receiver operating characteristic curve of 93.79\% and an accuracy of 91.3\%, which further improved with the inclusion of neuro-psychological scores. Beyond structural imaging, X Cui et al \cite{cui2018classification} and J Zhang et al \cite{zhang2017alzheimer} explored the fusion of structural imaging with advanced classification techniques. X Cui et al presented a Minimum Spanning Tree classification framework and their approach achieved remarkable classification accuracies with 98.3\%, 91.3\%, and 77.3\% for MCI vs. normal controls, AD vs. normal controls, and AD vs. MCI, respectively. Zhang et al relied on a landmark discovery algorithm and landmark-based affine registration to classify MCI based on structural imaging features, resulting in a classification accuracy of 88.03\%. \par
As successful as structural neuroimaging has been in early AD detection, functional neuroimaging has emerged as an efficient approach. Resting-state fMRI (rs-fMRI) has gained prominence due to its reduced dependence on individual cognitive abilities, making it a promising tool for the early identification of disease processes. The examination of rs-fMRI data offers valuable insights into functional alterations observed in individuals with MCI. Disruptions in functional connectivity networks are examined by analyzing data from resting-state fMRI. Biomarkers, utilizing changes in brain structure, function, and connectivity, have been devised to distinguish among healthy individuals, varying stages of MCI, and Alzheimer's disease.
Transitioning to functional magnetic resonance imaging (fMRI), Khazaee et al\cite{khazaee2017classification} explored a graph theoretical approach. Through a combination of advanced machine learning techniques and multivariate Granger causality analysis, their work evaluated resting-state fMRI data to calculate directed graph measures reflecting brain network connectivity. Impressively, they achieved an accuracy of 93.3\% in classifying AD, MCI, and healthy controls. This accentuated the potential of resting-state fMRI in capturing intricate differences in brain network patterns associated with MCI.
T Zhang et al\cite{zhang2019classification} highlighted the crucial role of careful frequency band selection and feature selection algorithms in differentiating MCI from healthy controls. The minimal redundancy maximal relevance (mRMR) algorithm, when applied to the slow-5 frequency band, emerged as the optimal choice, achieving an accuracy of 83.87\%. This reaffirmed the potential of fMRI and graph theory in elucidating the neural basis of MCI.
The deep fusion model proposed by Lyu et al\cite{lyu2021classification} integrated a cross-model deep network for multi-modal brain image data and a fully-connected neural network for gene expression data. Remarkably, the model learned a parameter representing the ratio of imaging to genetic features during the classification process. Drawing data from the Alzheimer's Disease Neuroimaging Initiative (ADNI), their study achieved an accuracy of 82.3\%, emphasizing the potential of deep learning in uncovering non-linear relationships between brain structure and function and complementing genetic data in the classification of AD/MCI patients.
Payan et al\cite{payan2015predicting} introduced a learning algorithm aimed at discriminating between healthy brains and diseased brains using MRI images as input. This approach involved deep artificial neural networks, specifically a combination of sparse autoencoders and convolutional neural networks. Notably, the application of 3D convolutions on the entire MRI image surpassed the performance of 2D convolutions on slices, resulting in a remarkable classification accuracy of 92.11\% when distinguishing MCI from healthy individuals.

Contrary to these works that encompass both structural and functional imaging modalities, in this paper, we study a very different approach that utilizes fMRI time series of six different brain networks to understand neurodegeneration in MCI. This is achieved by employing analysis at two extremely different scales. First, in order to examine the spatio-temporal characteristics of the fMRI time series, at each ROI at local scale, nonlinear characteristics are studied. Recurrence attributes are analyzed using the time series at each individual ROI, belonging to a particular brain network using recurrence plots. At the global scale, all ROIs comprising an entire brain network is studied using graph framework. We perform Eigen decomposition of graph adjacency matrix and the respective Eigen values and Eigen vectors are used as features for classification. Comparing these two approaches across all six brain networks, it is seen that the Eigen decomposition analysis performs better during classification for all six networks as compared to recurrence plots. The best classification between HC and MCI is obtained using Eigen vector of the highest Eigen value of the adjacency matrix of DMN, resulting in 89\% accuracy. However, using recurrence plots, classification accuracy is reduced to 77\% for DMN and maximum accuracy of 89.47\% is obtained for occipital network. It is also observed that using both approaches, the obtained classification accuracy is comparable in case of occipital, sesorimotor, cingulo-opercular and cerebellum brain networks. In this approach, we aim to understand the effects of neurodegeneration through classification, using two extreme scales of analysis. However, combining multiple scales could lead to better classification and understanding. Besides, devising disparity measures would reduce confusion in the classifier, leading to better performance. Several techniques could be leveraged to tackle the problem of  high dimensionality that is a bane with deep learning based approaches. Also, as an extension of the graph-based approach, the existing diverse tools in graph theory can be exploited for better performance. 

\section*{Conclusion}

The proposed work utilizes two extreme scales of analysis, one being global at the brain network scale and another being local, which studies each ROI in a brain network separately.  Graph-based framework analyzes brain networks and non-linear time series analysis studies dynamics at each ROI separately. By incorporating information-rich recurrence plots into a modified DenseNet121 deep learning network, we successfully classified subjects as healthy or with MCI. This dual-pronged  strategy provides a comprehensive understanding of brain dynamics, offering valuable insights into both local and global connectivity patterns. On fMRI volumes of 50 HC and 50 MCI subjects taken from publicly available ADNI dataset, we have been able to infer that certain ROIs show reduced activity in MCI, such as, PCC. Besides we observe greater predictability in time-series in MCI.


\section*{Proposed Methodology}
\label{sec:methodology}
\begin{figure}[ht]
\centering
\includegraphics[width=\linewidth]{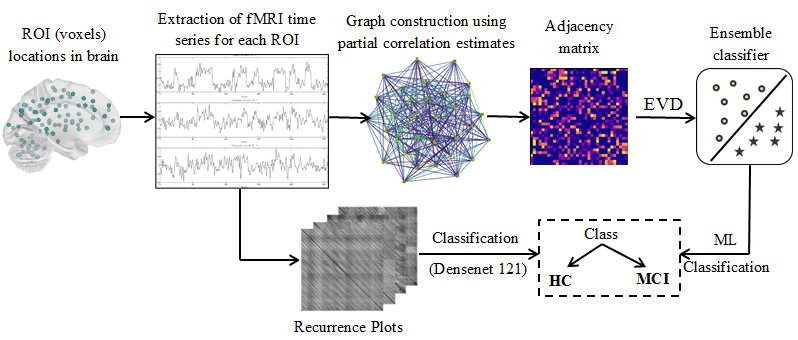}
\caption{Block schematic of proposed multi-scale methodlogy}
\label{fig:stream}
\end{figure}

The proposed methodology for distinguishing individuals with MCI from those with healthy cognitive function involves a sequence of critical steps. These stages encompass initial data preprocessing, the segmentation of the brain into networks, the generation of optimal time series representations, the extraction of relevant features, and, lastly, the classification process. A visual depiction of the proposed methodology is shown in Figure~\ref{fig:stream}. 

\subsection*{Extraction of time series from fMRI image volume}
\label{ssec:fmritots}
The process of selecting regions of interest (ROIs) played a pivotal role in our analysis of brain imaging data. We utilized the potential of Dosenbach's ROIs, a widely recognized atlas that provides precise coordinates for the structural divisions of the cerebral cortex\cite{dosenbach2010prediction}. This atlas is a standard point of reference in the realm of functional neuroimaging studies. These ROIs are organized into six distinct network types, encompassing cerebellar(18), opercular(32), default mode(34), frontoparietal(21), occipital(22), and sensorimotor(33) networks. These networks collectively encompass a multitude of interconnected brain regions. In our analysis, we utilized a total of 160 ROIs, each represented by a sphere with a 5 mm radius. Within each of these regions, multiple time series data associated with the constituent voxels are encapsulated. \par
Handling multiple time series from each brain region presented a significant challenge in our study. To tackle this difficulty, we devised a method to determine an optimal representation for the time series within each brain region. Instead of simply averaging all the voxels within a region, which might introduce outliers and inaccuracies, we adopted the concept of regional homogeneity (ReHo). ReHo has been successfully employed in prior studies to measure the temporal similarity of fMRI signals during resting states\cite{reho_cl, reho2}.
ReHo quantifies the degree of coherence and temporal similarity in the dynamics of a specific brain area. It is based on the premise that intrinsic brain activity is characterized by clusters of voxels working together, rather than isolated individual voxels. This approach enables us to capture patterns of local connectivity within the brain and, by extension, extract representative time series data that encapsulate the functional characteristics of each region.
To implement ReHo analysis, we utilized the Analysis of Functional NeuroImages (AFNI) software\cite{afni1,afni2}, which facilitated quantifying the similarity and coherence of fMRI signals within each region. ReHo analysis was performed at the voxel level, involving the calculation of the Kendall Coefficient of Concordance (KCC). This metric assessed the temporal similarity of a given voxel's time series with its 26 neighboring voxels in all directions, providing insight into the local temporal dynamics.
We selected candidate voxels from the ReHo maps within each brain region based on their level of temporal synchronization. The mean ReHo value within each region was employed as the threshold for selecting these candidate voxels. This approach ensured that the representation of each region accurately captured the characteristics of voxels that exhibited higher temporal homogeneity within their respective areas providing valuable insights into brain connectivity patterns.

\subsection*{Spatial encoding of fMRI time series using recurrence plot}

To understand the complexity in the time series pertaining to each of the ROI's we utilise non linear complexity analysis approach. We use the concept of Recurrence plots \cite{cao2002determining} which is very effective and widely accepted method to visualise and understand non linear time series. The concept of recurrences was formally introduced by Henri Poincaré in the year 1890. The Poincaré recurrence theorem states that certain dynamical systems will, after a sufficiently long but finite time, return to a state arbitrarily close to (for continuous state systems) or exactly the same as (for discrete state systems) their initial state. Thus, recurrence is a fundamental property of many dynamic systems and therefore various processes in nature \cite{myref1}. 
In 1987, Eckmann et al., \cite{eckmann1995recurrence} introduced the method of recurrence plots to visualize the recurrences of dynamical systems. 
Later, recurrence plots employ recurrence quantification analysis, which is a method of nonlinear data analysis quantifying the number and duration of recurrences of a dynamical system based on its state space trajectory \cite{marwan2007recurrence, webber2005recurrence} and commonly used to analyze time series generated from nonlinear deterministic systems, such as those found in the stock market and weather forecasting. A study by A. Fabretti and M. Ausloos \cite{fabretti2005recurrence} utilized recurrence plots and recurrence quantification analysis to detect the critical regime in financial indices. Another study by Peter Martey Addo et al. \cite{ADDO2013416} exploits the concept of recurrence plots to study the stock market to locate hidden patterns, non-stationarity, and to examine the nature of these plots in events of financial crisis. In particular, the recurrence plots are employed to detect and characterize financial cycles. 
The idea of recurrence plots has also been adopted in neuroscience domain. Bielski K et al. \cite{bielski2021parcellation} used recurrence quantification analysis to create Parcellation of the human amygdala. Kang Y et al .\cite{kang2023recurrence} employed  recurrence quantification analysis to measure the duration, predictability and complexity of the periodic processes of the nonlinear DMN time series to study Schizophrenia. The current work exploits patterns in the visualization images of recurrence plots and uses a CNN based approach to classify MCI vs. HC. 
In this study, we have showed the distinguishing pattern of recurrence plot between MCI and HC, which further aid to classify these two groups.

\subsubsection*{Construction of Recurrence Plots}

Using the extracted time series we have a balanced dataset of fMRI volume time series of 50 MCI and 50 HC.
Time series($t_s$) for a single ROI can be expressed as:
\begin{equation}
    t_s = \{v_1, v_2, \ldots, v_n\}
\end{equation}
Here, $v_i$ represents the voxel intensity of the ROI at instance $i$.

In order to perform non-linear analysis on the time series, the optimal values of the required parameters, embedding dimension ($M$), time lag ($\tau$),  and number of states ($K$), are determined using the traditional Cao's algorithm \cite{cao2002determining}. For time series $\{v_1, v_2, \ldots, v_n\}$ consisting of $N$ timestamps, we create $K$ state vectors. Each state vector $\overrightarrow{s_i}$ is an $M$-dimensional vector defined as, 

\begin{equation}
   \overrightarrow{s_i} = (v_i, v_{i+\tau}, v_{i+2\tau}, \ldots, v_{i+(N-1)\tau})
   \label{eq:1}
\end{equation}

\textbf{Data Matrix}

Data matrix, represented as $D$, consists of these states, stacked together, as shown in Equation. \ref{eq:2}. The dimension of $D$ is, $(K \times M)$. This matrix $D$ will now be utilized to define the recurrence matrix (RM).
    \begin{equation}
       D = 
       \begin{bmatrix}
       \overrightarrow{s_1}\\
       \overrightarrow{s_2}\\
       \vdots\\
       \overrightarrow{s_n}\\
        \end{bmatrix}_{(K \times M)}
        \label{eq:2}
    \end{equation}

\textbf{Recurrence matrix (RM)}

Distance between two state vectors is computed using their Euclidean distance, quantifying the dissimilarity between them. Hence, for a given tuple (i,j), the distance between states, $\overrightarrow{s_i}$ and $\overrightarrow{s_j}$, is the $(i,j)^{th}$ element of the RM, given by Eq \ref{eq:3},

\begin{equation}
\centering
    RM(i)(j) = dist(\overrightarrow{s_i}, \overrightarrow{s_j})
    \label{eq:3}
\end{equation}

Hence, the resulting RM is as defined below:

\begin{equation}
\centering
    RM = 
    \begin{bmatrix}
    dist(\overrightarrow{s_1},\overrightarrow{s_1}) & \cdots & dist(\overrightarrow{s_1},\overrightarrow{s_K}) \\
    dist(\overrightarrow{s_2},\overrightarrow{s_1}) & \cdots & dist(\overrightarrow{s_2},\overrightarrow{s_K}) \\
    \vdots & \vdots & \vdots\\
    dist(\overrightarrow{s_K},\overrightarrow{s_1}) & \cdots  & dist(\overrightarrow{s_K},\overrightarrow{s_K}) \\
    \end{bmatrix}_{(K \times K)}    
    \label{eq:4}
\end{equation}

\begin{figure}[htbp]
\centering
\includegraphics[width=\linewidth]{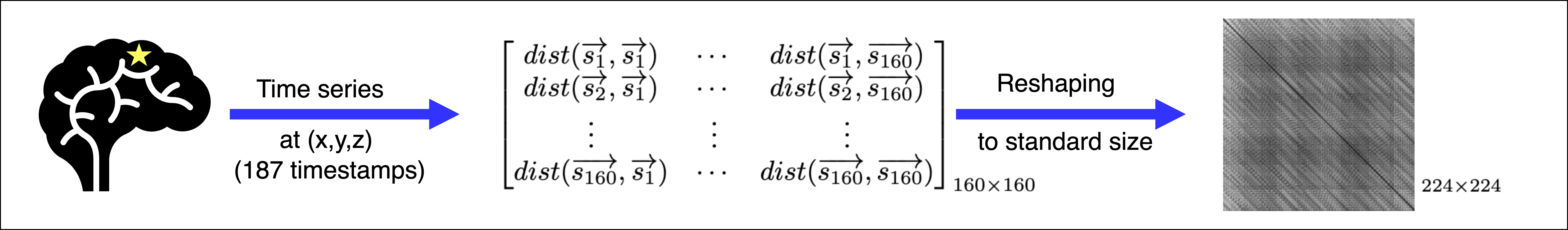}
\caption{Illustration of conversion of an fMRI time series to recurrence plots. Given a brain region, say, for instance post cingulate 108 belonging to DMN, we have a representative time series data of the BOLD signal from this region. This is converted to recurrence matrix of size $160\times160$ and then resized to uniformly-sized $224\times224$ to visualize the recurrence plot.  
}
\label{fig:recurrencematrix}
\end{figure}

The recurrence plot visualization shown in Figure~\ref{fig:recurrencematrix} is a gray-scale single channel image, capturing the evolution of the time series. Hence, the information at a specific ROI is now available as an image of size $1\times224\times224$. This leads to a high-dimensional representation of each brain network as a data point of n$\times224\times224$ ('n' is the number of ROIs).

\subsection*{Graph construction from fMRI time series}

Correlation analysis has long been a prominent technique in studying the connectivity between different brain regions\cite{ref1,ref16,ref17}. However, the commonly used Pearson correlation, primarily captures the marginal association between network nodes. This approach has limitations as it may not adequately capture direct or true connectivity between nodes. For instance, significant correlations between nodes A and B may arise due to their shared connections with a third node, C, even when A and B are not directly connected \cite{ref16,ref17}. Consequently, distinguishing between network edges reflecting true connectivity and those influenced by confounding factors becomes challenging. Addressing this issue, partial correlation has emerged as a valuable statistical technique \cite{ref16,ref17,ref18}. Partial correlation estimates the correlations after removing spurious effects from all other nodes in the network. This makes it a more accurate measure of network connectivity, with a zero value indicating an absence of direct connectivity between node pairs. Extensive evidence from the literature suggests that partial correlation stands out as one of the superior techniques, demonstrating high sensitivity in identifying true functional connectivity between network nodes \cite{ref16,ref17,ref19,ref43,ref44,ref45,ref46}. \par

Partial correlation of two time series between node `$i$' and node `$j$' at t is defined as the correlation between $T_{i}$ and $T_{j}$ conditioning on all the other nodes \cite{ref16}, i.e.:\\
\[ \rho_{ij} = corr(T_{i}, T_{j}\mid T_{ij}), \text{ where } T_{ij}=\{T_{k}: 1\leq k\leq n;\; k \neq i \neq j \}\]


The widely used approach for computing partial correlation utilize the inverse covariance matrix, commonly known as the precision matrix.
In our study, the undirected graphs are constructed using the partial correlations as edge weights. Each ROI is treated as a node in the graph, and the edges are determined by the partial correlation values between the corresponding ROIs. This method results in subject-specific graphs with weighted edges, reflecting the strength and nature of connectivity between brain regions.


\subsubsection*{Graph feature extraction}

The obtained graph using partial correlation is represented by a connectivity matrix (also known as adjacency matrix) which is a symmetric matrix of size $n \times n$, where, $n$ is the number of ROIs (nodes). Each $(i,j)$-th entry in this matrix denotes the strength of the edge connected between nodes `$i$' and `$j$'. Eigen decomposition is applied on each adjacency matrices to facilitate a deeper understanding of the graph's characteristics. The eigenvalues and eigenvectors associated with the largest eigenvalue are given as input features for our classification framework.

\subsubsection*{ROI analysis}
In addition to deriving the adjacency matrices for each subject's graph, corresponding degree matrices are also calculated. The degree matrix provides crucial information about the connectivity of nodes within the graph, specifically indicating the number of edges connected to each node. From each degree matrix we rank the ROIs (nodes) in descending order based on the number of edges connected to each node to identify the most highly connected nodes within a subject's graph. This will also paves the way for an in-depth examination of the roles played by these nodes in the context of HC and MCI subjects. 
This will not only provide valuable insights into the differences in network's organization between HC and MCI but also paves the way for an in-depth examination of the roles played by these ROIs while differentiating MCI from healthy individuals.

\subsubsection*{Classification}

In the context of detecting MCI from fMRI time series data, the selection of an appropriate machine learning learning and deep learning classifiers holds significant importance. This subsection explores various architectural choices for classifying MCI from HC. Densenet-121 neural network model and decision tree are employed for classifying HC Vs. MCI using two different types of features: (1) the 2D recurrence plots and (2) Eigen values and Eigen vectors, respectively. \par
Utilizing recurrence plots, we generate 2D images of size $224\times224$ for each ROI. This results in a tensor of dimensions $n \times 224 \times 224$, where '$n$' represents the total number of ROIs for a specific brain network. This is fed to Densenet-121 model. Densenet-121, recognized for its effectiveness in image classification tasks, is chosen for its unique architecture featuring densely connected blocks. These blocks promote efficient learning and feature reuse, making it well-suited for the intricacies of fMRI data. The model is trained using the cross-entropy loss function, a suitable choice for binary classification tasks. The Adam optimizer is employed with a learning rate of 0.001, and early stopping is used. The entire experiment is conducted using Kaggle notebooks and the PyTorch deep learning framework. Leveraging Kaggle's computational resources, particularly the $2\times 16$GB NVIDIA Tesla T4 GPU, ensures efficient model training and evaluation. \par
For classification using Eigen value decomposition, this study employs a classifier that integrates a base classifier utilizing a decision tree and the Bootstrap Aggregation technique \cite{dietterich2000experimental}. The ensemble classifier combines multiple weak classifiers to enhance overall performance compared to individual classifiers. Decision tree classifiers serve as the base classifiers within the ensemble. For this, classification framework is implemented in MATLAB 2019a on a laptop featuring a configuration with 16 GB RAM and an Intel Core i7-10750H processor (2.6 GHz). The default parameter values for the ensemble classifier using bagging technique are employed, with the number of cycles set to 400, determined through hyperparameter optimization. A 10-fold cross-validation procedure is executed.


\section*{Data availability}
The dataset analyzed during the current study is available on Alzheimer’s Disease Neuroimaging Initiative (ADNI) website (National Institutes of Health,
USA).

\bibliography{sample}


\section*{Author contributions statement}

All authors contributed to the study conception and design. A.A. and N.A conducted the experiment and analyzed the results under the guidance of N.S. at Centre for Brain Research, IISc Bangalore. Problem statement of the research was formulated by N.S. The data preparation and interpretation of data are performed by A.R. and A.A. and N.A. The draft of the manuscript is written by D.B., A.R., A.A. and N.A. and has been critically revised by N.S. for important intellectual content. All authors reviewed and approved the final manuscript.

\section*{Competing interests}
The authors declare no competing interests.

\section*{Additional information}
Correspondence and requests for materials should be addressed to D.B.

\end{document}